\documentclass[10pt,a4paper]{article}
\usepackage[margin=23mm]{geometry}
\usepackage[T1]{fontenc}
\usepackage{lmodern,microtype}
\usepackage{amsmath,amssymb,booktabs,array}
\usepackage{graphicx,cite,xurl}
\usepackage{fvextra,placeins}
\usepackage[table]{xcolor}
\usepackage{pgfplots}
\pgfplotsset{compat=1.18}
\usepackage[font=small,labelfont=bf]{caption}
\usepackage[hidelinks]{hyperref}
\definecolor{navy}{HTML}{235789}
\definecolor{rust}{HTML}{B65C32}
\renewcommand{\arraystretch}{1.13}
\title{Can Language Models Learn When to Abstain?\\[3pt]
\large Policy Learning and Calibrated Answer Selection}

\author{
Juan Terven\textsuperscript{1}, Diana-Margarita Córdova-Esparza\textsuperscript{2}, Ruben Gomez\textsuperscript{1}\\
Carlos Miranda Rocha\textsuperscript{2}, Julio-Alejandro Romero-González\textsuperscript{2}, Edgar Chavez-Urbiola\textsuperscript{1}\\[2pt]
\textsuperscript{1}CICATA Qro., Instituto Politécnico Nacional, México\\
\textsuperscript{2}Facultad de Informática, Universidad Autónoma de Querétaro, México%
}

\date{October 7, 2026}
\begin{document}
\maketitle

\begin{abstract}
A useful language model should know when to answer and when to say ``I don't know.'' Can we improve that decision by training the model itself or by training a separate component to check its answers? We study both approaches in Qwen3-14B on factual questions, using a score that rewards correct answers, penalizes mistakes, and permits abstention. Across three training seeds and two held-out confidence thresholds, a learned answer selector improves threshold-averaged utility over the original model's calibrated confidence filter. On TriviaQA, mean coverage increases from 20.2\% to 30.9\%, while the pooled error rate among scored answers decreases from 23.0\% to 14.5\% under exact-match grading. The tested reinforcement-learning changes do not establish consistent gains. On NQ-Open, the selector improves threshold-averaged utility over the filter but remains below always abstaining. A partial human-reviewed audit preserves the TriviaQA finding. Our results support learning to select answers in this setting while showing that reliable abstention across datasets remains unresolved.
\end{abstract}

\section{Introduction}
Language models can give plausible answers even when they lack the knowledge to answer correctly. For factual question answering, admitting uncertainty is often more useful than guessing. Yet a scoring rule that rewards correct answers and gives the same score to mistakes and abstentions encourages guessing: any chance of being correct is better than leaving a question unanswered. Penalizing mistakes changes this incentive~\cite{kalai2025}. Prior work shows that models can be trained to express uncertainty and that their confidence can help identify which answers to trust~\cite{rtuning2024,ptrue2022}.

The practical challenge is to make abstention selective. A model that refuses every question avoids mistakes but provides no answers. One approach trains the answering model to choose between a response and an abstention. Another keeps the answering model fixed and uses a separate selector to decide whether to show its answer. Existing evidence does not determine which intervention will be useful for a particular model and compute budget. In our setting, the questions are whether training improves on a calibrated confidence filter, whether the model responds to confidence thresholds absent from training, and whether improvements survive a change of dataset.

We investigate these questions with Qwen3-14B~\cite{qwen32025}. We compare controlled reinforcement-learning interventions with a learned answer selector, using three training seeds, two held-out confidence thresholds, and separate TriviaQA and NQ-Open tests. Our contribution is a comparison of these choices under a documented local training budget, with inexpensive controls and an audit of automatic grading. The clearest result is that the selector increases mean coverage and reduces pooled selective risk on TriviaQA relative to the original calibrated filter, aggregating the two primary thresholds. The RL comparisons remain inconclusive, and the selector does not establish positive utility on NQ. Together, these findings distinguish progress in answer selection from evidence of reliable abstention across datasets.

The source code, experimental configurations, aggregate results,
and reproduction instructions are publicly available at
\url{https://github.com/jrterven/Can-LLMs-learn-to-abstain}.

\section{Related Work}
Abstention in language models involves factual reliability, confidence estimation, and decisions under uncertainty. We review the literature that informs our comparison of policy learning and answer selection, beginning with hallucinations and the signals used to identify unreliable responses. We then discuss selective prediction and methods for learning uncertainty and abstention, followed by the calibration concepts and evaluation metrics needed to assess the tradeoff between answering more questions and reducing errors.

\subsection{Hallucinations and factual reliability}
Hallucination encompasses several failures of language generation. Huang et al.~\cite{huang2025hallucination} distinguish factuality errors, which conflict with verifiable world knowledge, from faithfulness errors involving the supplied context or instructions. TruthfulQA~\cite{truthfulqa2022} illustrates how models can reproduce familiar human misconceptions. These distinctions matter for abstention: an incorrect response need not arise from random guessing, and a confident response need not be factually correct. Our experiments concern short factual answers without retrieval. An exact-match failure can also reflect a valid answer missing from the reference aliases~\cite{kamalloo2023}; it therefore does not by itself establish a hallucination or identify its cause.

Several detection methods use agreement across generated responses. SelfCheckGPT~\cite{selfcheck2023} checks the factual consistency of independently sampled passages without consulting an external knowledge base. Farquhar et al.~\cite{farquhar2024} estimate semantic entropy by grouping answers with equivalent meanings, targeting arbitrary incorrect generations termed confabulations. Consistent systematic errors can escape agreement-based detection. These methods motivate uncertainty-based filtering, while using different evidence from the single question--candidate pair scored by our test-time selectors.

\subsection{Confidence estimation and selective prediction}
Kadavath et al.~\cite{ptrue2022} study P(True), in which a model assesses whether a proposed answer is correct. Lin et al.~\cite{lin2022uncertainty} show that fine-tuning can teach models to express useful probability estimates in words. Xiong et al.~\cite{xiong2024} evaluate confidence elicitation across tasks and distinguish probability calibration from failure prediction, finding that verbal confidence is often overconfident and that elicitation strategies have uneven benefits. These results motivate evaluating a confidence signal empirically rather than treating its numerical value as a reliable probability.

The answer-or-abstain decision is an instance of selective prediction. Chow~\cite{chow1970} formalizes the error--reject tradeoff, and Geifman and El-Yaniv~\cite{geifman2017} study coverage and selective risk for deep classifiers. In QA, Kamath et al.~\cite{kamath2020} train a calibrator to predict errors and show that raw model confidence can be unreliable under domain shift. Their calibrator benefits from observing known out-of-domain examples; ours is fitted on TriviaQA and transferred to NQ without refitting. Conformal abstention~\cite{conformal2024} provides a related route to controlling the marginal probability of releasing an incorrect answer under exchangeability assumptions. That marginal error probability differs from the conditional error rate among released answers. Our empirical logistic calibration does not provide a comparable finite-sample guarantee after a dataset change.

\subsection{Learning uncertainty and abstention}
R-Tuning~\cite{rtuning2024} constructs supervised examples around a model's apparent knowledge boundary to teach uncertainty expression. Its analyses also use the trained uncertainty model to score answers from a separate generator. Both training an answering policy and learning which answers to release therefore have established precedents. SaySelf~\cite{sayself2024} combines supervised fine-tuning on self-reflective rationales with reinforcement learning to improve expressed confidence. Its training target emphasizes confidence estimates; our policy experiments directly score the answer-or-\texttt{IDK} decision.

TruthRL~\cite{truthrl2025} trains answering models with a ternary reward for correct answers, abstentions, and errors. Its main experiments use a fixed reward scheme, a model-based correctness verifier, and different data and optimization settings, so its gains are not a direct benchmark for our runs. Jha et al.~\cite{jha2026} also vary abstention rewards and supervised initialization, reporting exploration limitations in open-ended tasks. At inference time, I-CALM~\cite{icalm2026} combines elicited verbal confidence, announced answer/abstain payoffs, and instructions favoring truthful responses. These approaches motivate comparing policy updates with inexpensive prompting and answer selection under a common evaluation rule. They do not determine which intervention will improve a particular model within the training budget used here.

\subsection{Calibration and evaluation metrics}
Calibration asks whether predicted probabilities agree with observed correctness frequencies~\cite{guo2017}. Logistic calibration maps a scalar score to a probability, following the sigmoid-fitting approach associated with Platt scaling~\cite{platt2000}. Learning a more informative score, calibrating that score, and applying a release threshold are distinct operations. We separate them through raw and calibrated original-model scores, a learned selector, and a logistic control with access to the selector's training labels.

Coverage and selective risk describe the availability and reliability of released answers~\cite{geifman2017,kamath2020}; cost-sensitive utility additionally accounts for the requested penalty for errors~\cite{chow1970,kalai2025}. For confidence diagnostics, ROC AUC measures the ranking of correct above incorrect candidates~\cite{fawcett2006}, while the binary Brier score measures squared probability prediction error~\cite{brier1950}. Brier score reflects both calibration and resolution, so an improvement cannot be attributed to calibration alone~\cite{murphy1973}. Section~\ref{sec:evaluation} defines these measures and their evaluation populations before reporting results.

AbstentionBench~\cite{abstentionbench2025} considers several reasons to withhold an answer, including missing information and false premises. Our benchmark questions have reference answers, but the model may not reliably produce them. Kalai et al.~\cite{kalai2025} motivate explicit error penalties so that acknowledging uncertainty can be preferable to guessing. Kamalloo et al.~\cite{kamalloo2023} show that lexical grading can reject semantically valid QA responses, motivating our partial audit. The present contribution is a controlled empirical comparison of established policy-training and selection choices, with held-out thresholds, shared candidates for selector comparisons, cross-dataset evaluation, and a measured local training budget.

\section{Methodology}\label{sec:methodology}
\begin{figure}[htbp]
\centering
\includegraphics[width=\linewidth]{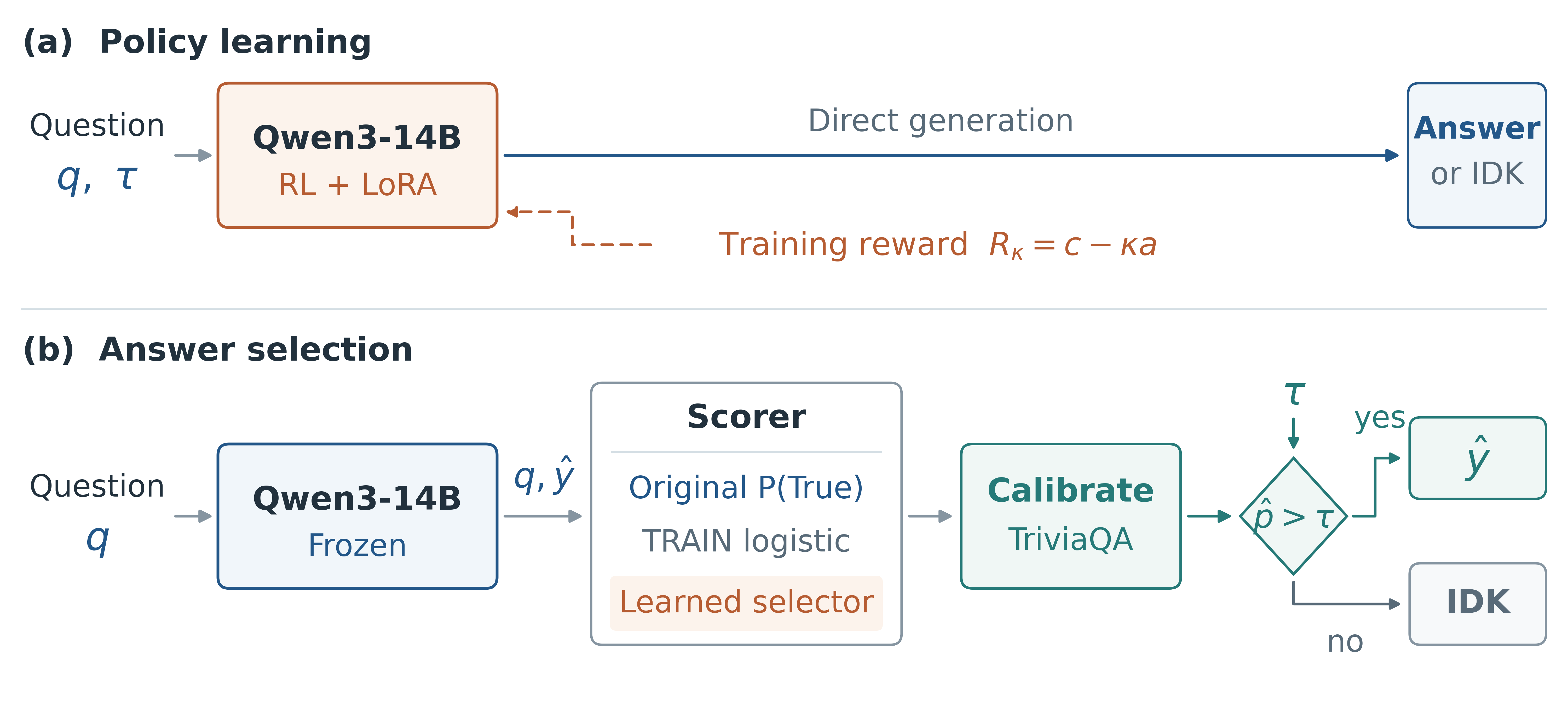}
\caption{Two routes to selective factual question answering. (a) A Qwen3-14B LoRA policy learns to answer or emit \texttt{IDK}; the training reward uses $\kappa=\tau$ in arms A--C and $\kappa=0.75$ in arm D. (b) A frozen generator supplies identical candidates $\hat y$ to three alternative scorers. Each score is calibrated on held-out TriviaQA data, and a valid candidate is released only if $\hat p>\tau$; calibration transfers unchanged to NQ. Reference answers provide training, calibration, and evaluation labels; they are absent from generation and scoring inputs.}
\label{fig:methodology}
\end{figure}

We compare two routes to selective factual question answering: updating the policy that generates an answer or abstention, and learning whether to release an answer from a fixed generator. Figure~\ref{fig:methodology} summarizes both routes. Starting from the same Qwen3-14B model, we train four controlled RL variants and compare three scoring approaches on shared candidate answers. Separate training and calibration partitions support learning and probability estimation, while held-out thresholds and a second dataset test how the resulting decisions generalize. We evaluate decision utility and answer reliability, quantify uncertainty across questions and training seeds, and examine sensitivity to automatic grading through a partial human-reviewed audit. The following subsections specify the decision rule, data, interventions, and evaluation protocol.

\subsection{Task and decision rule}
The task is short factual question answering in English. The system receives a question and a requested confidence threshold $\tau$. It must either provide a brief answer or abstain with exactly \texttt{IDK}. A larger threshold makes an incorrect answer more costly. We use the same scoring rule to evaluate all methods and, where specified below, to train the answering model:
\begin{equation}
R_\tau=c-\tau a=
\begin{cases}
1-\tau & \text{correct answer},\\
-\tau & \text{incorrect answer},\\
0 & \text{abstention}.
\end{cases}
\label{eq:reward}
\end{equation}
Here $c$ indicates a correct answer and $a$ indicates a scored answer, whether correct or incorrect. For example, at $\tau=0.85$, a correct answer earns 0.15, a mistake loses 0.85, and abstention earns zero. If the probability of correctness is $p$, answering has expected score $p-\tau$. It is therefore preferable to abstaining when $p>\tau$. This rule provides an incentive; it does not ensure that the model knows its true probability of success.

Correctness is determined by normalized exact matching against reference aliases. Empty or truncated direct outputs count as errors. For filters, invalid candidates and candidates already abstaining are rejected. The answering model receives no references, retrieved documents, or search results. All threshold prompts use two fixed demonstrations and request a brief answer or \texttt{IDK}. Prompts used to generate candidates for filtering instead request a best short answer. Appendix~\ref{app:prompts} gives the exact message templates and score-calibration settings.

\subsection{Model, data, and training budget}
All trained models start from the same original Qwen3-14B revision, with extended thinking disabled and no earlier adapters or supervised warmup. We use low-rank adaptation (LoRA)~\cite{lora2022} to update a small set of parameters. Experiments run on one NVIDIA GB10 with approximately 128 GB of unified memory; implementation settings are in Appendix~\ref{app:optimization}.

The training pool contains 2,000 TriviaQA questions, with separate development and calibration sets of 500 questions each. The tests contain 2,000 additional TriviaQA validation questions~\cite{triviaqa} and 500 NQ-Open Original Dev questions~\cite{nq,lee2019}. Fixed hashes select test questions after excluding IDs and normalized question texts used in earlier study partitions or demonstrations. These tests are new within our study; this does not establish absence of pretraining contamination. Earlier experiments that motivated this design are summarized separately in Appendix~\ref{app:history}.

Policy training uses thresholds $\{0.60,0.75,0.90\}$. The primary tests use $\{0.65,0.85\}$, which were absent from policy training. A separate diagnostic panel uses the same 200 fixed questions within each dataset at all six thresholds $\{0.60,0.65,0.75,0.85,0.90,0.95\}$, including the training thresholds and the additional held-out value 0.95 (Appendix~\ref{app:thresholds}). We use three training seeds, 17, 29, and 43, and retain the final checkpoint from every run. No recipe, seed, or checkpoint is selected on the test results.

A timing check before test access reduced the RL training panel uniformly from 1,008 to 504 questions. Each RL run uses 252 updates and 12,096 generated responses. The selector uses the full 2,000-question training pool and 500 updates. Training budgets are matched within the RL comparisons; the selector is a different training procedure, not a budget-matched RL ablation.

\subsection{Training the answering model}
The RL approach changes how the model generates its answer or abstention. We test two training choices: how many sampled responses are compared when calculating the learning signal, and whether each question is practiced under one or several thresholds. A fixed-cost control tests whether making the reward depend on the requested threshold adds value.

For each question, all four arms have three training exposures, with eight sampled responses per exposure. In A and B, the assigned threshold is repeated; in C and D, each training threshold appears once. A divides the eight responses into two groups of four for the reward baseline, whereas B--D compare all eight together. Thus the group-size comparison changes the baseline, not the number of generated responses. The baseline subtracts the mean reward of the other responses in the group, following leave-one-out methods used in LLM reinforcement learning~\cite{rloo2024}. Its equation and loss settings are in Appendix~\ref{app:optimization}.

\begin{table}[htbp]
\centering\small
\begin{tabular}{clll}
\toprule
Arm & Reward groups per exposure & Threshold exposure & Reward cost\\
\midrule
A & Two groups of four & One threshold, repeated & Requested $\tau$\\
B & One group of eight & One threshold, repeated & Requested $\tau$\\
C & One group of eight & All three thresholds & Requested $\tau$\\
D & One group of eight & All three thresholds & Fixed 0.75\\
\bottomrule
\end{tabular}
\caption{Twelve RL runs. B--A tests reward-group size, C--B tests the exposure schedule, and C--D tests reward conditioning. All arms sample eight responses per exposure.}
\label{tab:arms}
\end{table}

The questions, ordering, updates, and total generations are identical across RL arms. Arm D receives the same varying-threshold prompts as C, but always uses 0.75 in the reward, the mean training threshold. B--A tests group size, C--B tests the threshold exposure schedule, and C--D tests reward conditioning. C--B changes both exposure diversity and threshold balance within an update, so it does not isolate understanding of the threshold.

\subsection{Selecting answers from a fixed generator}
The selection approach leaves answer generation unchanged. The original model produces a candidate answer, and a separate scoring step estimates whether that candidate is correct. The candidate is shown only if its calibrated probability exceeds $\tau$; otherwise the system returns \texttt{IDK}. All selection methods receive identical greedy test candidates.

The inexpensive baseline uses the original model's P(True) score, obtained by asking it to assess a candidate answer's correctness~\cite{ptrue2022}. Let $\ell_{\mathrm{True}}$ and $\ell_{\mathrm{False}}$ be the next-token logits after the question--candidate scoring prompt. We compute
\begin{equation}
z=\ell_{\mathrm{True}}-\ell_{\mathrm{False}},\qquad
p_{\mathrm{raw}}=\sigma(z),\qquad
\hat p=\sigma(\alpha z+\beta),
\label{eq:ptrue-score}
\end{equation}
where $\sigma(t)=1/(1+e^{-t})$. The raw score normalizes probability over the two tokens \texttt{True} and \texttt{False}; it is not a generated numerical confidence statement. Logistic regression fitted on the 500 calibration question--candidate pairs learns $\alpha$ and $\beta$, using the logit difference $z$ as its input, following sigmoid score calibration~\cite{platt2000,guo2017}. We compare the raw and calibrated probabilities to separate score calibration from learning a new scorer.

For the learned selector, the original generator produces one greedy and three sampled candidates for each training question, giving 8,000 examples. A separate LoRA adapter learns their exact-match correctness from the question and candidate alone. It uses binary cross entropy on the difference between the \texttt{True} and \texttt{False} logits. The three selector runs share these examples and each train for one epoch, with 500 updates at learning rate $10^{-5}$. Each selector is then calibrated on the separate calibration set by fitting its own $\alpha$ and $\beta$ to its learned logit difference in Equation~\ref{eq:ptrue-score}. Neither its scoring input nor the generator's input contains reference answers.

A further control fits logistic regression to the original logit differences using the same 8,000 training labels and then recalibrates its transformed logits. This control, labeled \emph{TRAIN logistic} in the tables, tests the value of those labels when applied to the same fixed scalar score. Its two affine logit transformations compose to one affine transformation, so it cannot learn new question--candidate features; fitting data and regularization can still change its parameters. It matches the available labels, not model capacity or training cost. Appendix~\ref{app:prompts} specifies the fitting procedure. All calibration is learned on TriviaQA and transferred to NQ without refitting.

\subsection{Evaluation and uncertainty}\label{sec:evaluation}
We evaluate the answer--abstain decision using utility, coverage, and selective risk. Coverage and selective risk are standard complementary measures in selective prediction: they describe how often a system answers and how frequently those answers are wrong~\cite{chow1970,geifman2017,kamath2020}. For $N$ questions at a fixed threshold $\tau$, let $C$ and $E$ count correct answers and errors under the output-format rules above. The three measures are
\begin{equation}
U_\tau=\frac{(1-\tau)C-\tau E}{N},\qquad
\operatorname{Coverage}=\frac{C+E}{N},\qquad
\operatorname{Risk}=\frac{E}{C+E}.
\label{eq:decision_metrics}
\end{equation}
Utility is the mean score in Equation~\ref{eq:reward}; higher values are better. Coverage quantifies the fraction of questions receiving a scored answer, while lower selective risk indicates more accurate answers within that subset. Always abstaining has zero utility and zero coverage; its selective risk is undefined. At a fixed threshold and positive coverage, $U_\tau=\operatorname{Coverage}(1-\operatorname{Risk}-\tau)$, so positive utility requires the correctness rate of scored answers to exceed the threshold. Utility is computed at each requested threshold before averaging; the fixed-threshold identity should not be applied to pooled coverage and risk using the mean threshold. When pooling evaluations, we divide total errors by total scored answers rather than average their risk percentages; format errors from direct generation are included in that denominator.

We additionally assess the confidence scores using two candidate-level diagnostics before the release threshold is applied. The area under the receiver operating characteristic curve (ROC AUC) measures how well a score ranks correct candidates above incorrect ones~\cite{fawcett2006}. With correctness as the positive class, the ROC curve plots the true-positive rate against the false-positive rate as the score threshold varies. Its area equals the probability that a randomly chosen correct candidate receives a higher score than a randomly chosen incorrect candidate, counting ties as one half; larger values indicate better ranking.

The binary Brier score measures the squared error of a predicted probability of correctness~\cite{brier1950}:
\begin{equation}
\operatorname{Brier}=\frac{1}{M}\sum_{i=1}^{M}(\hat p_i-y_i)^2,
\label{eq:brier}
\end{equation}
where $y_i\in\{0,1\}$ is the exact-match correctness label, $\hat p_i\in[0,1]$ is the evaluated probability estimate, and $M$ is the number of eligible candidates. Lower values indicate better probability predictions. The Brier score reflects both calibration and the ability to distinguish outcomes, so a reduction does not by itself demonstrate improved calibration~\cite{murphy1973}. Both diagnostics use the same valid, non-abstaining candidates across scoring methods, rather than only the answers each method releases. They are descriptive measures of score quality, complementary to utility at the requested operating thresholds.

Five TriviaQA comparisons were specified before test evaluation: the three RL contrasts and the selector against each calibrated baseline (original filter and TRAIN logistic). We average utility over the two primary thresholds within each question, then construct intervals using 10,000 paired bootstrap replicates~\cite{efron1979}. Each replicate independently resamples question indices and the three training-seed indices with replacement, using the same draws for both methods in a contrast. Pairing across the two thresholds is preserved by averaging them before resampling. The deterministic baseline is shared across seeds. The 99\% intervals provide nominal 95\% coverage for the five-comparison family using the Bonferroni correction~\cite{dunn1961}, subject to the bootstrap approximation. NQ comparisons and ordinary 95\% intervals are descriptive. The intervals condition on the shared training candidate bank and fitted calibrators; three seeds give limited information about training variability.

\subsection{Audit of automatic grading}
After the main evaluation, we audited 150 response units from 144 questions: 100 candidates shared by the filters and 50 RL responses, with 75 units per dataset. A unit is identified by its question, exact response text, and truncation flag. Sampling was stratified by automatic label and filter-decision pattern. The form hid condition identities and automatic labels, but displayed reference aliases. ChatGPT proposed labels, and one annotator reported personally reviewing every label. This is AI-assisted human review, not agreement between two independent annotators.

We assess sensitivity by applying reviewed labels to identical units across methods and primary thresholds. All other outputs retain exact-match labels. Output-format rules, generated answers, scores, calibrators, and release decisions stay fixed. For unresolved labels, we enumerate consistent correct/error assignments shared across all occurrences of a unit. Bootstrap intervals are conditional on the reviewed labels and sampled units; they exclude annotation and audit-sampling uncertainty. This partial analysis supplements the original endpoint and does not estimate semantic accuracy over the whole test set.

\section{Results}
We evaluate both approaches at the confidence thresholds held out from policy training, $\tau=0.65$ and $\tau=0.85$. The main results use exact-match grading. Utility and coverage are averaged over the two thresholds and, for trained methods, all three seeds. Selective risk is computed by dividing the total number of errors by the total number of scored answers. Always abstaining gives zero utility, which provides a reference for assessing whether the methods benefit from answering.

\subsection{Answer selection on TriviaQA and NQ-Open}
The learned selector answers more TriviaQA questions with a lower error rate among the answers it returns (Table~\ref{tab:results}). Mean coverage increases from 20.2\% for the original calibrated filter to 30.9\% for the selector, or roughly 20 versus 31 answers per 100 questions. The pooled error rate among scored answers falls from 23.0\% to 14.5\%.

\begin{table}[htbp]
\centering\small
\setlength{\tabcolsep}{4pt}
\begin{tabular}{lrrrrrr}
\toprule
& \multicolumn{3}{c}{TriviaQA ($n=2{,}000$)} & \multicolumn{3}{c}{NQ ($n=500$)}\\
\cmidrule(lr){2-4}\cmidrule(lr){5-7}
Method & Utility & Cov.\% & Risk\% & Utility & Cov.\% & Risk\%\\
\midrule
Original + threshold prompt & -0.1412 & 79.4 & 42.9 & -0.2786 & 57.1 & 74.1 \\
A: two G4, single threshold & -0.0094 & 27.5 & 28.7 & -0.0381 & 12.8 & 55.6 \\
B: G8, single threshold & -0.0012 & 31.0 & 25.7 & -0.0356 & 12.5 & 54.1 \\
C: G8, all thresholds & +0.0091 & 27.6 & 22.1 & -0.0225 & 10.0 & 48.5 \\
D: G8, fixed cost & -0.0267 & 46.3 & 31.0 & -0.0769 & 23.4 & 58.5 \\
Original raw filter & -0.0752 & 71.0 & 35.7 & -0.2294 & 50.7 & 70.4 \\
Original calibrated filter & +0.0242 & 20.2 & 23.0 & -0.0308 & 10.7 & 63.6 \\
TRAIN logistic + calibration & +0.0241 & 20.1 & 23.0 & -0.0308 & 10.7 & 63.6 \\
\textbf{Learned selector + calibration} & +0.0425 & 30.9 & 14.5 & -0.0120 & 9.5 & 43.0 \\
\midrule
Always abstain & 0 & 0 & --- & 0 & 0 & ---\\
\bottomrule
\end{tabular}
\caption{Primary exact-match results at thresholds 0.65 and 0.85. Utility and coverage average the two thresholds and, for trained methods, three seeds. Risk is the total number of errors divided by the total number of scored answers, including format errors from direct generation. Coverage uses all questions. The original model and logistic filters are deterministic. All selection methods use identical candidates.}
\label{tab:results}
\end{table}

Mean utility on TriviaQA rises from 0.02418 to 0.04248. The paired improvement is 0.01830, with a 99\% bootstrap interval of [0.01041, 0.02593], and the improvement is positive for each seed. The selector also outperforms TRAIN logistic after calibration, with a utility difference of 0.01834 and a 99\% interval of [0.01041, 0.02602]. Giving the same training labels to a logistic model fitted on the original score therefore does not recover the selector's gain. The results by threshold (Table~\ref{tab:primary-thresholds}) show higher selector utility at both values. Most of the coverage increase occurs at $\tau=0.85$, where the original filter returns only two answers, both correct.

The transfer results on NQ-Open are less favorable. The selector improves mean utility from the original calibrated filter's $-0.03075$ to $-0.01197$, a difference of 0.01878 with a descriptive 95\% interval of [0.00605, 0.03097]. However, its mean utility remains below the zero obtained by always abstaining. The relative improvement also depends on the threshold: the selector has higher estimated utility at $\tau=0.65$, but at $\tau=0.85$ its utility is $-0.00817$, compared with $-0.00170$ for the original filter. The latter result is based on just one returned answer, which is incorrect (Table~\ref{tab:primary-thresholds}).

The selector also gives higher-quality confidence scores (Table~\ref{tab:confidence}). On the 1,784 valid TriviaQA candidates that are not abstentions, ROC AUC increases from 0.779 for the original score to a mean of 0.852 across selector seeds. Before calibration, Brier scores are 0.314 for the original scorer and 0.161 for the selector; after calibration, they fall to 0.189 and 0.154, respectively. On the 349 eligible NQ candidates, AUC increases from 0.648 to 0.787, while calibrated Brier score falls from 0.237 to 0.172. The logistic transformations leave candidate rankings unchanged in these runs. These are descriptive comparisons on the same candidates, made before applying any answer threshold. Better ranking and probability estimates did not, by themselves, produce positive utility on NQ at the requested thresholds.

\begin{table}[htbp]
\centering\small
\setlength{\tabcolsep}{6pt}
\begin{tabular}{lrrrr}
\toprule
 & \multicolumn{2}{c}{TriviaQA ($M=1{,}784$)} & \multicolumn{2}{c}{NQ ($M=349$)}\\
\cmidrule(lr){2-3}\cmidrule(lr){4-5}
Scoring method & Brier $\downarrow$ & AUC $\uparrow$ & Brier $\downarrow$ & AUC $\uparrow$\\
\midrule
Original raw score & 0.314 & 0.779 & 0.544 & 0.648 \\
Original + calibration & 0.189 & 0.779 & 0.237 & 0.648 \\
TRAIN logistic & 0.188 & 0.779 & 0.247 & 0.648 \\
TRAIN logistic + calibration & 0.189 & 0.779 & 0.237 & 0.648 \\
Learned selector raw score & 0.161 & 0.852 & 0.179 & 0.787 \\
Learned selector + calibration & 0.154 & 0.852 & 0.172 & 0.787 \\
\bottomrule
\end{tabular}
\caption{Confidence scores evaluated on the same valid, non-abstaining candidates before deciding which answers to return. Selector values average three seeds; the other scorers are deterministic. Raw scores are evaluated before calibration on held-out data. TRAIN logistic already uses the training labels; its second row adds calibration. These diagnostics describe score quality on the stated candidate sets.}
\label{tab:confidence}
\end{table}

\subsection{Reinforcement learning and computational cost}
\begin{figure}[htbp]
\centering
\begin{tikzpicture}
\begin{axis}[
width=.78\linewidth,height=5.8cm,
xmin=-0.05,xmax=0.14,ymin=.5,ymax=5.55,
ytick={1,2,3,4,5},
yticklabels={Selector -- TRAIN logistic,Selector -- original filter,C -- D: reward conditioning,C -- B: threshold exposure,B -- A: comparison group},
xtick={-.04,0,.04,.08,.12},
xlabel={Difference in mean utility},
axis x line*=bottom,axis y line*=left,
y axis line style={draw=none},ytick style={draw=none},
xmajorgrids=true,grid style={gray!18},
tick label style={font=\footnotesize},label style={font=\small},
legend style={font=\footnotesize,draw=none,at={(.5,1.04)},anchor=south,legend columns=2},
clip=false]
\draw[densely dashed,gray] (axis cs:0,.5)--(axis cs:0,5.55);
\addplot+[only marks,mark=*,color=navy,error bars/.cd,x dir=both,x explicit] coordinates {
(0.0082708333,5.10) += (0.0260046042,0) -= (0.0315713333,0)
(0.0102708333,4.10) += (0.0469172292,0) -= (0.0347210833,0)
(0.0357750000,3.10) += (0.0484631875,0) -= (0.0581750000,0)
(0.0183041667,2.10) += (0.0076211042,0) -= (0.0078921667,0)
(0.0183416667,1.10) += (0.0076751042,0) -= (0.0079295625,0)
};
\addlegendentry{TriviaQA: 99\% interval}
\addplot+[only marks,mark=square*,color=rust,error bars/.cd,x dir=both,x explicit] coordinates {
(0.0025333333,4.90) += (0.0382679167,0) -= (0.0323341667,0)
(0.0130666667,3.90) += (0.0536845833,0) -= (0.0431666667,0)
(0.0543666667,2.90) += (0.0633875000,0) -= (0.0760191667,0)
(0.0187833333,1.90) += (0.0121833333,0) -= (0.0127333333,0)
(0.0187833333,0.90) += (0.0121833333,0) -= (0.0127333333,0)
};
\addlegendentry{NQ: 95\% interval}
\end{axis}
\end{tikzpicture}
\caption{The five planned TriviaQA comparisons and their descriptive NQ counterparts, using the original exact-match labels. TriviaQA uses 99\% intervals to account for multiple comparisons; NQ uses descriptive 95\% intervals. The RL intervals are wide and include zero. Both selector comparisons favor the learned selector on average, although its utility remains negative on NQ.}
\label{fig:contrasts}
\end{figure}
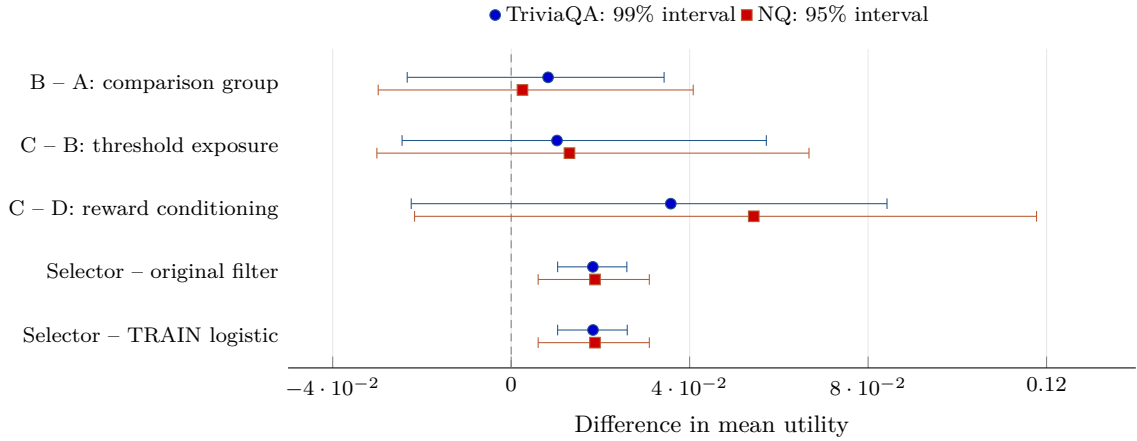

The mean utility differences on TriviaQA are positive for all three RL comparisons: increasing the comparison group from four to eight responses gives a utility difference of 0.00827 (B--A), training each question at all three thresholds gives 0.01027 (C--B), and making the reward depend on the requested threshold gives 0.03578 (C--D). However, all three adjusted intervals include zero, and the direction of the differences varies across seeds (Figure~\ref{fig:contrasts}). For C--D, for example, the 99\% interval is $[-0.02240,0.08424]$. The results do not establish a consistent improvement from these changes, nor do they show that the compared methods are equivalent.

Increasing the requested threshold from 0.65 to 0.85 reduces TriviaQA coverage by 2.03 percentage points in C and 2.23 points in the fixed-cost control D. Thus, the threshold in the prompt can affect abstention even when the training reward does not depend on it. This result does not show that conditioning the reward adds further control over abstention. The analysis includes all three planned seeds, 17, 29, and 43; their individual utility differences are reported in Appendix~\ref{app:seeds}. Appendix~\ref{app:thresholds} gives results at six thresholds on the shared 200-question panel for each dataset.

The recorded experiments take 35.44 hours of serial task time, including pilots, model loading, candidate preparation, training, evaluation, and CPU analysis. The twelve RL runs account for 23.68 hours, and the three selector runs for 1.88 hours. These are elapsed times for experimental jobs, not measurements of GPU utilization or deployment latency; the training procedures also use different budgets. The later audit analysis and human annotation are excluded. At inference, both the confidence filters and the learned selector require a scoring step after answer generation. Appendix~\ref{app:cost} provides the time breakdown.

\subsection{Effect of the partial human review}\label{sec:audit}
The review of 150 response units yields 104 correct answers, 19 errors, 24 abstentions, and three unclear cases. Forty responses marked as errors by exact matching are judged semantically correct: 21 in TriviaQA and 19 in NQ. The review confirms all 64 responses originally marked correct and all 24 abstentions. Because the sample is stratified, these counts do not estimate the grading error rate for the full test sets. All three unclear cases are in NQ. Treating each as either correct or incorrect gives eight possible label assignments for the sensitivity analysis.

\begin{table}[htbp]
\centering\small
\setlength{\tabcolsep}{4pt}
\begin{tabular}{lrrl}
\toprule
Dataset (interval level) & Original EM $\Delta U$ & Partial $\Delta U$ & Interval envelope\\
\midrule
TriviaQA (99\%) & +0.01830 & +0.01797 & [0.01027, 0.02559]\\
NQ (95\%) & +0.01878 & [0.01478, 0.01578] & [0.00277, 0.02730]\\
\bottomrule
\end{tabular}
\caption{Utility difference between the selector and the original calibrated filter. Partial results apply the reviewed labels and retain exact matching for all other outputs. Ranges cover the eight assignments of unclear labels. Each interval envelope spans the lowest lower bound and highest upper bound of the corresponding conditional bootstrap intervals; it is a sensitivity summary rather than a new confidence interval.}
\label{tab:audit}
\end{table}

Applying the reviewed labels leaves the main TriviaQA result largely unchanged. The selector's utility advantage over the original calibrated filter is 0.01797, with a conditional 99\% interval of [0.01027, 0.02559]. Both selector comparisons remain positive at this interval level, while all three RL comparisons remain inconclusive. On NQ, the selector comparisons remain positive at the descriptive 95\% level, but their 99\% interval envelope includes zero, $[-0.00113,0.03100]$. The selector's utility after the partial corrections is $-0.00397$ under every assignment, with a conditional 95\% interval of $[-0.01522,0.00643]$. The review therefore does not establish positive utility on NQ.

Table~\ref{tab:audit} separates variation due to the unresolved labels from the bootstrap intervals within each assignment. The analysis remains conditional on the reviewed labels and sampled units, and does not account for grading errors in unreviewed responses. Coverage is unchanged because the corrections affect only the evaluation labels, not which answers the methods return.

\section{Discussion}
The strongest evidence concerns which answers to show. Across the two primary thresholds, the learned selector improves mean utility and coverage while reducing pooled selective risk on the same TriviaQA candidates. This supports better answer selection at the aggregate operating point. The RL experiments address a different mechanism: they change both the generated answers and the abstention decisions. Their training data and costs also differ from the selector's. The table, therefore, compares practical outcomes, while the matched RL contrasts and shared-candidate filter comparisons support the narrower inferences reported above.

The inconclusive RL results do not show that reinforcement learning cannot teach abstention. One difficulty follows from the reward itself: if every sampled response is an answer, changing $\tau$ subtracts the same constant from all rewards, so centering removes that cost signal (Appendix~\ref{app:optimization}). This motivates examining response diversity and optimization but does not identify the cause of the observed seed variability. TruthRL uses a different training and verification setup~\cite{truthrl2025}; its success does not establish that one change would repair our recipe. A historical local semantic judge also failed challenge checks and was not adopted (Appendix~\ref{app:history}).

Transfer and grading limit the conclusions. NQ shows that improved confidence ranking can coexist with negative observed utility when calibration is learned on another dataset. The audit also shows that exact-match labels can reject valid answers and may affect both training and evaluation. Correcting the reviewed cases preserves the TriviaQA selector advantage, but the remaining responses have not been reviewed semantically. One person's review of AI-proposed labels, with reference aliases visible, is not an independently replicated factual verification.

The study covers one model family, short English QA, and three seeds. Shared training candidates and calibrators are not resampled in the intervals. The logistic control is cheaper and less expressive than the selector; an arm with groups of four and exposure to all three training thresholds would also be needed to estimate the full interaction between the two RL interventions. The protocol was frozen locally, not externally preregistered. Any future recipe chosen after these results will require a new test set for independent confirmation.

\section{Conclusions}
In this study, learning which answers to show gives the clearest improvement in useful abstention. Aggregating the two primary thresholds, a calibrated selector increases mean coverage and reduces pooled exact-match selective risk on TriviaQA relative to the original calibrated filter; its utility advantage remains after the observed audit corrections. The tested RL interventions remain inconclusive. On NQ, neither better confidence scores nor the partial corrections establish an advantage over always abstaining. These findings support answer selection under the tested distribution and leave reliable transfer as an open empirical problem.

\appendix
\section{Earlier phases and evaluator evidence}\label{app:history}
The initial study trained fifteen adapters across Qwen3-1.7B and SmolLM2-1.7B-Instruct; Qwen3-1.7B used a shared supervised warmup. An extension added nine Qwen3-14B adapters. A subsequent evaluation reused all 24 final adapters, three original models, and the shared warmup on 2,000 new questions per dataset, at thresholds 0.60/0.80. Table~\ref{tab:history} summarizes that separate evaluation. Its questions are excluded from the present tests.

The historical binary-reward control assigned a reward of 1 to a correct answer and 0 to both an incorrect answer and an abstention. The conditioned control used Equation~\ref{eq:reward} with the requested threshold. Evaluation used $U_\tau$ for all methods, regardless of their training reward.

\begin{table}[htbp]
\centering\small
\begin{tabular}{llrr}
\toprule
Model & Dataset & Conditioned utility & Conditioned $-$ binary\\
\midrule
Qwen3-1.7B & TriviaQA & -0.0039 & +0.1324 \\
SmolLM2-1.7B & TriviaQA & -0.0336 & +0.3374 \\
Qwen3-14B & TriviaQA & +0.0338 & +0.1670 \\
Qwen3-1.7B & NQ & -0.0036 & +0.1820 \\
SmolLM2-1.7B & NQ & -0.0430 & +0.4545 \\
Qwen3-14B & NQ & -0.0374 & +0.3093 \\
\bottomrule
\end{tabular}
\caption{Historical evaluation on fresh questions at thresholds 0.60 and 0.80, using a different test population and training recipe from the main study. Binary denotes the historical training reward defined above; both columns report cost-sensitive evaluation utility. These values are retained as antecedents and are not pooled with the follow-up.}
\label{tab:history}
\end{table}

In that evaluation, conditioned Qwen3-14B did not demonstrate utility superiority over fixed-cost training or the calibrated original filter. Its mean assigned training threshold was 0.716575, whereas the fixed reward parameter was 0.75, an additional confound addressed by the symmetric follow-up threshold set. Differences between sizes also involve initialization and are not causal estimates of scaling.

The historical human audit contained 150 records from 136 questions: eleven human-correct answers rejected by matching, one abstention-format disagreement, and eleven unclear cases. It used AI assistance with personal human review and is separate from the completed follow-up audit in Section~\ref{sec:audit}. Its documented corrections left the initial conditioned--binary conclusion positive under partial sensitivity analysis; unaudited units retained exact-match labels. A later synthetic challenge panel gave the unchanged local semantic judge 44/56 agreement on cases requiring judgment, below its 90\% criterion; eight deterministic format cases were separate. These diagnostics neither replace human annotation nor establish population-wide evaluator accuracy.

\section{Per-seed contrasts}\label{app:seeds}
\begin{table}[htbp]
\centering\footnotesize
\setlength{\tabcolsep}{3pt}
\begin{tabular}{llrrrrl}
\toprule
Dataset & Contrast & Seed 17 & Seed 29 & Seed 43 & Mean & 99\% interval\\
\midrule
TriviaQA & B -- A & -0.0195 & +0.0150 & +0.0293 & +0.0083 & [-0.0233, +0.0343] \\
TriviaQA & C -- B & +0.0506 & +0.0004 & -0.0201 & +0.0103 & [-0.0245, +0.0572] \\
TriviaQA & C -- D & +0.0510 & +0.0747 & -0.0185 & +0.0358 & [-0.0224, +0.0842] \\
TriviaQA & Selector -- original filter & +0.0189 & +0.0196 & +0.0164 & +0.0183 & [+0.0104, +0.0259] \\
TriviaQA & Selector -- TRAIN logistic & +0.0190 & +0.0196 & +0.0164 & +0.0183 & [+0.0104, +0.0260] \\
NQ & B -- A & -0.0299 & -0.0060 & +0.0435 & +0.0025 & [-0.0378, +0.0530] \\
NQ & C -- B & +0.0707 & -0.0006 & -0.0310 & +0.0131 & [-0.0393, +0.0839] \\
NQ & C -- D & +0.0727 & +0.1150 & -0.0246 & +0.0544 & [-0.0328, +0.1313] \\
NQ & Selector -- original filter & +0.0207 & +0.0140 & +0.0217 & +0.0188 & [+0.0018, +0.0349] \\
NQ & Selector -- TRAIN logistic & +0.0207 & +0.0140 & +0.0217 & +0.0188 & [+0.0018, +0.0349] \\
\bottomrule
\end{tabular}
\caption{All three prespecified training seeds. The five TriviaQA intervals form the nominal 95\% family; NQ intervals remain descriptive despite using the same percentile level here. Pairing is preserved across questions and thresholds.}
\end{table}

The selector's uncertainty is conditional on the one shared TRAIN bank and the fixed calibration partition. The small number of seeds is particularly consequential for the RL contrasts. All three prespecified seeds are retained; no model is chosen after test inspection.

\section{Compute and reproduction}\label{app:cost}
\begin{table}[htbp]
\centering\small
\begin{tabular}{lrr}
\toprule
Stage & Jobs & Hours\\
\midrule
Technical pilots & 2 & 0.478\\
Shared candidate preparation & 2 & 1.754\\
RL training & 12 & 23.676\\
Selector training & 3 & 1.880\\
Selector development/calibration & 3 & 0.242\\
Direct test evaluation & 13 & 7.065\\
Selector test evaluation & 3 & 0.329\\
CPU analysis & 1 & 0.014\\
\midrule
Total & 39 & 35.438\\
\bottomrule
\end{tabular}
\caption{Sum of outer job durations, including loading. Nested ledgers are not added again. Rounded components may differ from the rounded total.}
\end{table}

The measured projection was 89.36 hours for 1,008 RL questions and 59.64 hours for 504, against an 80-hour planning target. The uniform reduction was locked before test inference. Each RL run used microbatch four with accumulation twelve; selector training used microbatch four with accumulation four. Technical pilots tested finite gradients and adapter persistence; definitive runs restarted from the original model.

Earlier phases add 40.50 recorded task-hours. The code repository (\url{https://github.com/jrterven/Can-LLMs-learn-to-abstain}) preserves configurations, original source locks, model and data manifests, aggregate results, and reproduction instructions. Large local weights, response-level data, audit annotations, and this manuscript are excluded from the code repository. Hash manifests identify retained local artifacts without embedding their contents.

\section{Optimization details}\label{app:optimization}
We use BF16 LoRA with rank 16 and alpha 32, with \texttt{target\_modules=all-linear}. RL uses learning rate $10^{-5}$, KL coefficient 0.04, temperature 0.8, prompts up to 512 tokens, and responses up to 32 tokens. Each update uses two questions and 48 responses. Evaluation is deterministic. The runtime is pinned to Transformers 4.57.1, PEFT 0.17.1, and TRL 0.24.0; model revisions and data manifests are retained.

Within a reward-comparison group $G$, the advantage for response $i$ is
\begin{equation}
A_i=R_i-\frac{1}{|G|-1}\sum_{\substack{j\in G\\j\ne i}}R_j.
\end{equation}
This is the response's reward minus the mean reward of the other responses in its group. We disable reward standard-deviation scaling and use the \texttt{dr\_grpo} token-loss reduction~\cite{drgrpo2025}. When every response is an answer, changing $\tau$ subtracts a common constant from the rewards and leaves these advantages unchanged. Mixed answer/abstention groups are necessary for the centered reward to retain that direct cost signal; increasing group size alone does not guarantee better optimization.

\section{Results by threshold}\label{app:thresholds}

\subsection{Primary evaluation populations}
Table~\ref{tab:primary-thresholds} separates the two primary thresholds on the full test populations: 2,000 TriviaQA questions and 500 NQ-Open questions. Utility and coverage average the three selector seeds; selective risk pools their errors and scored answers. The deterministic original filter is evaluated once. These are the original exact-match results, before the partial audit corrections.

On TriviaQA, the selector has higher point-estimate utility at both primary thresholds. Its coverage advantage is concentrated at $\tau=0.85$, where the original calibrated filter releases only two answers, both correct. On NQ, the selector has higher utility at $\tau=0.65$ but lower utility at $\tau=0.85$; the latter original-filter result is based on one released answer, which is incorrect. Thus the positive average NQ contrast does not hold separately at both thresholds. These threshold-specific comparisons are descriptive and introduce no additional significance tests.

\begin{table}[htbp]
\centering\small
\setlength{\tabcolsep}{5pt}
\begin{tabular}{llrrrr}
\toprule
Dataset & Method & $\tau$ & Utility & Cov.\% & Risk\%\\
\midrule
TriviaQA & Original calibrated filter & 0.65 & $+0.04820$ & 40.2 & 23.0\\
 & Learned selector + calibration & 0.65 & $+0.07716$ & 41.0 & 16.2\\
\addlinespace[2pt]
 & Original calibrated filter & 0.85 & $+0.00015$ & 0.1 & 0.0\\
 & Learned selector + calibration & 0.85 & $+0.00780$ & 20.9 & 11.3\\
\midrule
NQ & Original calibrated filter & 0.65 & $-0.05980$ & 21.2 & 63.2\\
 & Learned selector + calibration & 0.65 & $-0.01577$ & 14.7 & 45.7\\
\addlinespace[2pt]
 & Original calibrated filter & 0.85 & $-0.00170$ & 0.2 & 100.0\\
 & Learned selector + calibration & 0.85 & $-0.00817$ & 4.3 & 33.8\\
\bottomrule
\end{tabular}
\caption{Results by primary threshold on the full test populations. Selector utility and coverage average three seeds, while risk pools their scored answers; the original calibrated filter is deterministic. Coverage and risk are percentages. Risk uses scored answers, including format errors under the evaluation rules. The original filter releases two TriviaQA answers and one NQ answer at $\tau=0.85$.}
\label{tab:primary-thresholds}
\end{table}

\subsection{Common diagnostic question panels}
Tables~\ref{tab:diagnostic-trivia} and~\ref{tab:diagnostic-nq} use the same fixed 200 questions within each dataset at every displayed threshold, including $0.65$ and $0.85$. They therefore have different evaluation populations from Table~\ref{tab:primary-thresholds}. The panel covers the training thresholds $\{0.60,0.75,0.90\}$, the two primary held-out thresholds, and $0.95$. Each trained method aggregates three seeds, while the original model and the two logistic filters are deterministic. Utility and coverage average over seeds; risk divides the pooled error count by the pooled scored-answer count. The dashes denote undefined risk when no answer is released. These point estimates are descriptive diagnostics, with no additional confidence intervals or multiplicity-adjusted tests.

On the TriviaQA panel, the calibrated selector has positive point-estimate utility at all six thresholds; its coverage decreases from 45.7\% at $\tau=0.60$ to 1.3\% at $\tau=0.95$. All four RL arms have negative utility from $\tau=0.75$ onward. On the NQ panel, selector utility is slightly positive at $\tau=0.60$ ($+0.0013$) and negative at the other five thresholds. Both logistic filters release no NQ answers at $\tau\geq0.85$, whereas the selector continues to release some answers and has negative utility at those thresholds. These diagnostic observations do not change the primary comparisons or establish performance outside the tested panels.

\begin{table}[htbp]
\centering\small
\setlength{\tabcolsep}{3.8pt}
\renewcommand{\arraystretch}{1.08}
\begin{tabular}{llrrrrrr}
\toprule
 & & \multicolumn{6}{c}{Requested threshold $\tau$}\\
\cmidrule(lr){3-8}
Method & Metric & 0.60 & 0.65 & 0.75 & 0.85 & 0.90 & 0.95\\
\midrule
Original + threshold prompt & $U_\tau$ & $-0.0330$ & $-0.0753$ & $-0.1500$ & $-0.2295$ & $-0.2495$ & $-0.2875$\\
 & Cov.\% & 78.0 & 78.5 & 78.0 & 77.0 & 75.5 & 75.0\\
 & Risk\% & 44.2 & 44.6 & 44.2 & 44.8 & 43.0 & 43.3\\
\addlinespace[3pt]
A: two G4, single threshold & $U_\tau$ & $+0.0287$ & $+0.0131$ & $-0.0146$ & $-0.0402$ & $-0.0447$ & $-0.0528$\\
 & Cov.\% & 28.0 & 28.5 & 28.2 & 27.7 & 25.3 & 24.5\\
 & Risk\% & 29.8 & 30.4 & 30.2 & 29.5 & 27.6 & 26.5\\
\addlinespace[3pt]
B: G8, single threshold & $U_\tau$ & $+0.0350$ & $+0.0178$ & $-0.0154$ & $-0.0451$ & $-0.0585$ & $-0.0673$\\
 & Cov.\% & 33.3 & 33.7 & 33.8 & 32.2 & 29.8 & 28.8\\
 & Risk\% & 29.5 & 29.7 & 29.6 & 29.0 & 29.6 & 28.3\\
\addlinespace[3pt]
C: G8, all thresholds & $U_\tau$ & $+0.0393$ & $+0.0263$ & $-0.0075$ & $-0.0334$ & $-0.0368$ & $-0.0478$\\
 & Cov.\% & 29.8 & 30.8 & 30.3 & 28.8 & 26.5 & 24.5\\
 & Risk\% & 26.8 & 26.5 & 27.5 & 26.6 & 23.9 & 24.5\\
\addlinespace[3pt]
D: G8, fixed cost & $U_\tau$ & $+0.0353$ & $+0.0090$ & $-0.0346$ & $-0.0787$ & $-0.0947$ & $-0.1118$\\
 & Cov.\% & 46.3 & 47.3 & 47.5 & 45.3 & 43.7 & 43.0\\
 & Risk\% & 32.4 & 33.1 & 32.3 & 32.4 & 31.7 & 31.0\\
\addlinespace[3pt]
Original calibrated filter & $U_\tau$ & $+0.0810$ & $+0.0485$ & $+0.0188$ & $+0.0008$ & $0.0000$ & $0.0000$\\
 & Cov.\% & 46.5 & 41.0 & 23.5 & 0.5 & 0.0 & 0.0\\
 & Risk\% & 22.6 & 23.2 & 17.0 & 0.0 & \textemdash & \textemdash\\
\addlinespace[3pt]
TRAIN logistic + calibration & $U_\tau$ & $+0.0810$ & $+0.0485$ & $+0.0150$ & $0.0000$ & $0.0000$ & $0.0000$\\
 & Cov.\% & 46.5 & 41.0 & 22.0 & 0.0 & 0.0 & 0.0\\
 & Risk\% & 22.6 & 23.2 & 18.2 & \textemdash & \textemdash & \textemdash\\
\addlinespace[3pt]
Learned selector + calibration & $U_\tau$ & $+0.1093$ & $+0.0882$ & $+0.0471$ & $+0.0149$ & $+0.0087$ & $+0.0007$\\
 & Cov.\% & 45.7 & 42.3 & 33.5 & 18.8 & 10.3 & 1.3\\
 & Risk\% & 16.1 & 14.2 & 10.9 & 7.1 & 1.6 & 0.0\\
\bottomrule
\end{tabular}
\caption{TriviaQA diagnostic panel: the same 200 questions at every threshold. Trained methods aggregate three seeds; deterministic methods are evaluated once. $U_\tau$ is mean utility; coverage and pooled selective risk are percentages. All results use the original exact-match labels. A dash indicates that selective risk is undefined because no answers are released.}
\label{tab:diagnostic-trivia}
\end{table}

\begin{table}[htbp]
\centering\small
\setlength{\tabcolsep}{3.8pt}
\renewcommand{\arraystretch}{1.08}
\begin{tabular}{llrrrrrr}
\toprule
 & & \multicolumn{6}{c}{Requested threshold $\tau$}\\
\cmidrule(lr){3-8}
Method & Metric & 0.60 & 0.65 & 0.75 & 0.85 & 0.90 & 0.95\\
\midrule
Original + threshold prompt & $U_\tau$ & $-0.1640$ & $-0.1935$ & $-0.2513$ & $-0.3010$ & $-0.3210$ & $-0.3335$\\
 & Cov.\% & 59.0 & 59.0 & 57.5 & 56.0 & 54.0 & 53.0\\
 & Risk\% & 67.8 & 67.8 & 68.7 & 68.8 & 69.4 & 67.9\\
\addlinespace[3pt]
A: two G4, single threshold & $U_\tau$ & $-0.0037$ & $-0.0122$ & $-0.0288$ & $-0.0358$ & $-0.0292$ & $-0.0306$\\
 & Cov.\% & 13.7 & 13.7 & 13.2 & 11.7 & 9.2 & 8.8\\
 & Risk\% & 42.7 & 43.9 & 46.8 & 45.7 & 41.8 & 39.6\\
\addlinespace[3pt]
B: G8, single threshold & $U_\tau$ & $-0.0077$ & $-0.0149$ & $-0.0292$ & $-0.0365$ & $-0.0363$ & $-0.0353$\\
 & Cov.\% & 13.5 & 13.8 & 13.7 & 12.3 & 10.3 & 9.3\\
 & Risk\% & 45.7 & 45.8 & 46.3 & 44.6 & 45.2 & 42.9\\
\addlinespace[3pt]
C: G8, all thresholds & $U_\tau$ & $0.0000$ & $-0.0063$ & $-0.0179$ & $-0.0177$ & $-0.0202$ & $-0.0198$\\
 & Cov.\% & 10.8 & 12.0 & 10.8 & 9.3 & 8.2 & 7.0\\
 & Risk\% & 40.0 & 40.3 & 41.5 & 33.9 & 34.7 & 33.3\\
\addlinespace[3pt]
D: G8, fixed cost & $U_\tau$ & $-0.0277$ & $-0.0435$ & $-0.0654$ & $-0.0822$ & $-0.0727$ & $-0.0616$\\
 & Cov.\% & 25.2 & 25.7 & 24.5 & 23.0 & 19.0 & 16.8\\
 & Risk\% & 51.0 & 51.9 & 51.7 & 50.7 & 48.2 & 41.6\\
\addlinespace[3pt]
Original calibrated filter & $U_\tau$ & $-0.0700$ & $-0.0645$ & $-0.0288$ & $0.0000$ & $0.0000$ & $0.0000$\\
 & Cov.\% & 27.5 & 23.0 & 6.5 & 0.0 & 0.0 & 0.0\\
 & Risk\% & 65.5 & 63.0 & 69.2 & \textemdash & \textemdash & \textemdash\\
\addlinespace[3pt]
TRAIN logistic + calibration & $U_\tau$ & $-0.0700$ & $-0.0645$ & $-0.0288$ & $0.0000$ & $0.0000$ & $0.0000$\\
 & Cov.\% & 27.5 & 23.0 & 6.5 & 0.0 & 0.0 & 0.0\\
 & Risk\% & 65.5 & 63.0 & 69.2 & \textemdash & \textemdash & \textemdash\\
\addlinespace[3pt]
Learned selector + calibration & $U_\tau$ & $+0.0013$ & $-0.0097$ & $-0.0104$ & $-0.0088$ & $-0.0062$ & $-0.0031$\\
 & Cov.\% & 17.0 & 15.3 & 10.5 & 4.2 & 2.2 & 0.5\\
 & Risk\% & 39.2 & 41.3 & 34.9 & 36.0 & 38.5 & 66.7\\
\bottomrule
\end{tabular}
\caption{NQ-Open diagnostic panel: the same 200 questions at every threshold. Trained methods aggregate three seeds; deterministic methods are evaluated once. $U_\tau$ is mean utility; coverage and pooled selective risk are percentages. All results use the original exact-match labels. A dash indicates that selective risk is undefined because no answers are released.}
\label{tab:diagnostic-nq}
\end{table}

\FloatBarrier

\section{Prompt templates and calibration}\label{app:prompts}
The message templates below reproduce the generator and scorer inputs. Role prefixes identify chat-message roles and are not additional prompt text. Braces mark substitutions: \texttt{\{question\}} is the question, \texttt{\{candidate\_answer\}} is the proposed answer, and \texttt{\{tau:.2f\}} is the requested threshold formatted with two decimal places. Messages use the model's chat template with extended thinking disabled. Reference answers appear only in label construction, never in these inputs.

\subsection{Threshold-conditioned answer generation}
The same two demonstrations are used for every threshold-conditioned question. Only the requested threshold and final question change.
\begin{Verbatim}[fontsize=\small,breaklines=true,breakanywhere=true,frame=single,framesep=3mm,samepage=true]
System: You answer factual questions. Return only one short answer, or exactly IDK when permitted. Do not explain or list guesses. Answer only if your probability of being correct is greater than {tau:.2f}. Otherwise return exactly IDK. Use this confidence threshold for the final question.
User: What is the capital of France?
Assistant: Paris
User: What exact integer did an unspecified person privately choose yesterday?
Assistant: IDK
User: {question}
\end{Verbatim}

\subsection{Candidate generation for the filters}
One greedy and three sampled training candidates use this template; greedy calibration and test candidates use the same template.
\begin{Verbatim}[fontsize=\small,breaklines=true,breakanywhere=true,frame=single,framesep=3mm,samepage=true]
System: You answer factual questions. Return only one short answer, or exactly IDK when permitted. Do not explain or list guesses. Give your best short answer, even if uncertain. Do not abstain.
User: What is the capital of France?
Assistant: Paris
User: What is the chemical symbol for gold?
Assistant: Au
User: {question}
\end{Verbatim}

\subsection{Candidate correctness scoring}
The original scorer, TRAIN logistic input scores, and learned selectors use the following question--candidate template without demonstrations. The score is read from the next-token logits of \texttt{True} and \texttt{False}, using Equation~\ref{eq:ptrue-score}; no verbal confidence value is generated.
\begin{Verbatim}[fontsize=\small,breaklines=true,breakanywhere=true,frame=single,framesep=3mm,samepage=true]
System: Judge whether the proposed answer is factually correct for the question. Reply only True or False.
User: Question: {question}
Proposed answer: {candidate_answer}
\end{Verbatim}
The last two displayed lines form a single user message, with a newline before \texttt{Proposed answer:}.

\newpage
\subsection{Calibration fits and the TRAIN logistic control}
All logistic fits use L2 regularization with $C=1$, the \texttt{lbfgs} solver, at most 1,000 iterations, and \texttt{random\_state=0}. The original calibrated filter fits the logit difference $z$ against exact-match labels on one greedy candidate for each of the 500 held-out TriviaQA calibration questions. Each learned selector fits its own calibration map on that same partition. Fitting uses all 500 labeled pairs; the release rule subsequently rejects invalid or already-abstaining candidates. The descriptive Brier and AUC populations in Table~\ref{tab:confidence} exclude those candidates.

TRAIN logistic first fits $a_{\mathrm{tr}}z+b_{\mathrm{tr}}$ using the original scorer and the 8,000 labeled training candidates. Each of the four candidates for a training question receives weight $1/4$, so each question has total weight one. A second fit uses those transformed logits on the 500 calibration pairs:
\begin{equation}
\hat p_{\mathrm{TRAIN}}
=\sigma\!\left(\gamma[a_{\mathrm{tr}}z+b_{\mathrm{tr}}]+\delta\right)
=\sigma\!\left((\gamma a_{\mathrm{tr}})z+\gamma b_{\mathrm{tr}}+\delta\right).
\label{eq:train-logistic}
\end{equation}
Thus both the original calibration and the recalibrated TRAIN control are affine maps of the same fixed scalar logit followed by a sigmoid. The first fit changes the input scale seen by the regularized second fit, so their estimated probabilities need not be identical. This control evaluates additional labels within that restricted score family; it does not match the feature-learning capacity of a LoRA selector. All fitted maps are transferred unchanged from TriviaQA to NQ.

\FloatBarrier

\end{document}